%% file: main.tex
\pdfoutput=1
\documentclass[sigconf]{acmart}
\AtBeginDocument{%
  }

\setcopyright{acmlicensed}
\copyrightyear{2027}
\acmYear{2027}
\acmDOI{XXXXXXX.XXXXXXX}
\acmConference[WWW '27]{the ACM Web Conference 2027}{May 10--14, 2027}{Dublin, Ireland}
\acmBooktitle{The ACM Web Conference 2027 (WWW '27), Under Review}
\acmISBN{978-1-4503-XXXX-X/2018/06}

\usepackage{algorithm}
\usepackage{algpseudocode}
\usepackage{multirow}
\usepackage{xspace}
\usepackage{pifont}
\usepackage{enumitem}
\usepackage{subcaption}
\usepackage{tcolorbox}
\tcbuselibrary{breakable,skins}

\RequirePackage{xcolor}
\usepackage{colortbl}
\newcommand{\grouprow}[2]{\rowcolor{gray!12}\multicolumn{#1}{l}{\textit{#2}} \\}
\RequirePackage{etoolbox}

\makeatletter
\apptocmd{\@secfont}{\color{black}}{}{}
\apptocmd{\@subsecfont}{\color{black}}{}{}
\apptocmd{\@subsubsecfont}{\color{black}}{}{}
\apptocmd{\@parfont}{\color{black}}{}{}
\makeatother

\input{dfn}

\begin{document}
\addtolength{\textfloatsep}{-5pt}
\addtolength{\dbltextfloatsep}{-5pt}
\captionsetup{skip=5pt}

\title{Not Yet in the Profile: Uncertainty-Aware Dynamic User Profiling by Learning from Future Interactions}

\author{Jieyong Kim}
\affiliation{%
  \institution{University of Illinois, Urbana-Champaign}
  \city{Champaign}
  \country{USA}}
\email{jieyong2@illinois.edu}

\author{Neil Shah}
\affiliation{%
  \institution{Pinterest}
  \city{Bellevue}
  \country{USA}}
\email{neilshah@pinterest.com}

\author{Dong Wang}
\addtocounter{footnote}{1}
\makeatletter\g@addto@macro\@authornotes{\stepcounter{footnote}}\makeatother
\authornote{Corresponding author.}
\affiliation{%
  \institution{University of Illinois, Urbana-Champaign}
  \city{Champaign}
  \country{USA}}
\email{dwang24@illinois.edu}

\begin{abstract}
\input{sections/00abstract}
\end{abstract}



\maketitle

\makeatletter
\global\@topnum\@ne
\makeatother
\begin{figure}[!t]
  \centering
  \includegraphics[width=\columnwidth]{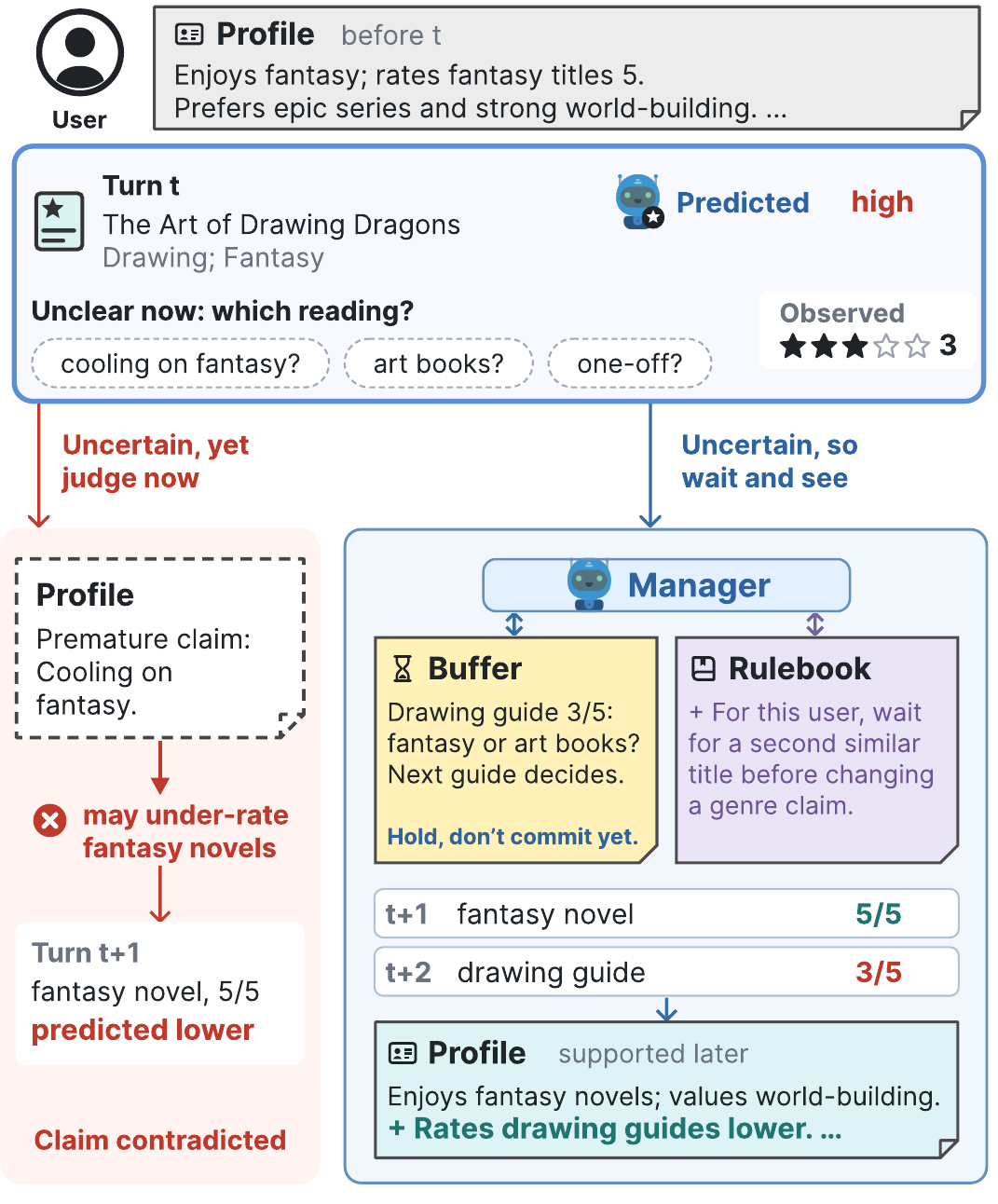}
  \caption{%
A new interaction can often be read in more than one way when it arrives.
Committing one reading at once can leave a wrong claim in the Profile that misleads later predictions, whereas holding it back limits this contamination.
}
  \Description{A fantasy fan rates a fantasy drawing guide 3/5. Committing at once claims the user is cooling on fantasy, so a later fantasy novel rated 5/5 is predicted lower. Holding the reading in the Buffer, under a user rule to wait for a second similar title, lets a later drawing guide rated 3/5 support that drawing guides are rated lower.}
  \label{fig:example}
\end{figure}

\section{Introduction}
\label{sec:introduction}
\input{sections/10introduction}

\section{Related Work}
\label{sec:related-work}
\input{sections/20relatedwork}

\section{Methodology}
\label{sec:methodology}

\input{sections/30methodology}

\section{Experiments}
\label{sec:experiments}

\input{sections/40experiments}

\section{Conclusion}
\label{sec:conclusion}
\input{sections/50conclusion}

\bibliographystyle{ACM-Reference-Format}
\bibliography{references}

\appendix

\vspace{-2pt}
\section{Limitations}
\label{sec:limitations}
\input{sections/71limitations}

\vspace{-2pt}
\section{Ethical Considerations}
\label{sec:ethical-considerations}
\input{sections/60ethical}

\vspace{-2pt}
\section{Implementation Details}
\label{app:implementation-details}
\input{sections/73implementation}

\vspace{-2pt}
\section{Baselines}
\label{app:baselines}
\label{app:implementation-details-adapted-baselines}
\input{sections/74baselines}

\vspace{-2pt}
\section{Datasets}
\label{app:datasets}
\input{sections/75datasets}

\vspace{-2pt}
\section{Ambiguous Events}
\label{app:ambiguous-events}
\input{sections/76ambiguous}

\vspace{-2pt}
\section{Additional Analyses}
\label{app:additional-analyses}
\input{sections/77analyses}

\end{document}

%% file: dfn.tex
\newcommand{\methodfull}{Uncertainty-aware Adaptive Profile Manager\xspace}
\newcommand{\method}{\textsf{UAPM}\xspace}
\newcommand{\trainingmethodfull}{Privileged On-Policy Distillation with Dual Specialization\xspace}
\newcommand{\trainingmethod}{POPD-DS\xspace}
\newcommand{\studentdistill}{Student Distillation\xspace}
\newcommand{\teacherdistill}{Future-informed Distillation\xspace}

%% file: sections/00abstract.tex
On the Web, user profiling builds compact user representations from interaction histories for efficient personalization, and dynamic user profiling updates them as users' behavior and preferences change.
However, new user behavior can be ambiguous, and its interpretation differs across users.
To address this, we introduce \methodfull (\method),
which separates the claims in a user profile from interpretations of new behavior that are not yet confirmed, through a Profile, a Buffer, and a Rulebook.
\method records user-specific interpretation and update rules in the Rulebook,
keeps ambiguous evidence in the Buffer outside the Profile as evidence for later decisions, and writes supported general and situational characteristics to the Profile.
We train \method with \trainingmethodfull (\trainingmethod), which learns from recorded future interactions during training without access to them at deployment.
Its first stage, Dual Specialization, distills a stronger model into a current-only Student and a future-informed model that becomes the Teacher.
In the second stage, privileged on-policy distillation (POPD), only the fixed Teacher receives future interactions and guides the Student's own management decisions alongside task rewards.
Experiments on three Web personalization tasks show that \method outperforms multiple baselines under the same downstream input budget, improving on the best same-budget baseline by up to 5\%.

%% file: sections/10introduction.tex
User profiling supports efficient personalization by summarizing a user's long interaction history into a compact representation. Recent approaches write it with large language models (LLMs) as natural-language profiles~\cite{kim2025exp3rt,wu2025rlpf} that LLMs read for personalized tasks.
Because a profile constructed once cannot reflect later changes in a user's behavior or preferences, \textit{dynamic user profiling}~\cite{bang2025pure,noh2025triple,chen-etal-2025-deeper} updates it as new interactions arrive.
On the Web, such profiles personalize e-commerce, review, and community question-answering services and serve as long-term user memory for LLM agents~\cite{zhong2024memorybank,chhikara2025mem0,xu2025amem,zhou2026mem1}.
Web services update them online from sparse, delayed, and incomplete feedback~\cite{zhao2026recap}, such as a single review, tag, or question, so misread interactions accumulate and keep shaping later recommendations and answers.
Written in natural language, they also make personalization explainable and correctable by users, so keeping them accurate as each user's behavior changes is a core problem of user modeling and personalization on the Web.

However, new interactions can be ambiguous, so updating a profile from currently available evidence is not always reliable.
For example, in Figure~\ref{fig:example}, a fantasy fan's 3/5 rating for a fantasy drawing guide can lead to the premature conclusion that the user is cooling on fantasy, whereas subsequent ratings of 5/5 for a fantasy novel and 3/5 for another drawing guide show that the user rates drawing guides lower.
Such ambiguous signals, which may reflect a lasting change, a one-time trial, or a preference that holds only for a specific kind of item, are not rare across personalization datasets.
For example, in book rating prediction~\cite{hou2024amazon}, a user sometimes gives a low rating to a genre they usually rate highly, and such ratings make up 6.3\% of all ratings.
After 64\% of these low ratings, the user rates the same genre 4 or higher again within the next ten interactions, compared with 78\% after a high rating.
This recovery rate also differs by user. Among the 833 sampled Amazon Books users with at least two such ratings (Appendix~\ref{app:ambiguous-events}), the middle half recover after 50\% to 80\% of them, a spread partly due to the few such ratings per user.
Likewise, in movie tagging~\cite{salemi2023lamp} and community question answering~\cite{salemi-zamani-2025-lamp}, 23\% of tags and 12\% of questions, respectively, involve a tag or a question community new to the user.
Of these new tags and communities, 63\% and 55\% reappear later while 37\% and 45\% do not, compared with 77\% and 88\% for previously used ones.
Committing every such signal can add claims that later turns do not support, while ignoring them misses real changes.
Existing profiling approaches can draw such premature conclusions because of two challenges.
First, \textbf{Uncertain New Behavior}: when an interaction arrives, whether it reflects a lasting change or a one-off exception is often not yet known, yet such methods must either commit an interpretation to the profile or discard it, and a wrong committed claim misleads subsequent predictions.
Second, \textbf{User-Dependent Commitment Criteria}: how much evidence is enough to commit an interpretation differs across users, as two consistent interactions may suffice for one user while three may not be enough for another, yet existing profiling methods keep no user-specific rules for when to commit, defer, or revise an interpretation.

To address these challenges, we introduce \methodfull (\method), which adapts how each user's new behavior is interpreted and used to update the profile by separating the claims in the Profile from interpretations of new behavior that are not yet confirmed.
\method does not estimate how likely each interpretation is, but decides how the Profile is managed while interpretations remain open.
At each turn, \method reads its previous state, the current behavior, and the downstream task model's predictions and feedback, and revises three components.
(1) The \emph{Profile} stores sufficiently supported user characteristics, marking them as \textit{situational} under specific conditions or \textit{general} across conditions.
(2) The \emph{Buffer} addresses Uncertain New Behavior: it keeps unresolved interpretations with their supporting or conflicting evidence outside the Profile, where the downstream task model does not read them but the manager can use them as evidence for later decisions.
(3) The \emph{Rulebook} addresses User-Dependent Commitment Criteria: it adapts user-specific rules, reflected from earlier management and prediction feedback, for managing the Buffer and revising Profile entries or their scope.
To the best of our knowledge, we are the first to introduce a profile manager that both holds unresolved interpretations of each interaction in a Buffer outside the Profile and keeps per-user rules for committing them.

Learning what to commit to the Profile requires supervision for when to commit, defer, or reconsider user information.
A reward on the updated Profile's next-turn predictions misses benefits that appear only several turns later.
Scoring later turns with the current Profile does not solve this, because the Profile is revised at every turn.
Predictions several turns later depend on the edits made in between, which one fixed Profile does not reflect.
Future interactions recorded in the training streams can instead show which current decisions later behavior supports.
In \textit{privileged on-policy distillation} (POPD)~\cite{penaloza2026pidistill,zhao2026opsd}, a Teacher that sees these future interactions guides a Student that manages profiles from current and past information, alongside task reward.
The guidance is \textit{on-policy}: the Teacher provides next-token distributions at the decision positions of the Student's own management trajectories.

However, existing privileged distillation approaches~\cite{zhao2026opsd,penaloza2026pidistill,tang2026psd} share a backbone between the Student and Teacher, so one model must both decide from current information and guide with future context, and such shared variants performed worse in our experiments.
We therefore propose \trainingmethodfull (\trainingmethod), which develops the two capabilities separately before on-policy training.
In the first stage, \textit{Dual Specialization}, knowledge distillation~\cite{hinton2015distilling} from a stronger model trains a current-only Student and a future-informed model, which becomes the Teacher in POPD.
In the second stage, \textit{POPD}, this fixed Teacher guides the Student with the remaining future interactions.
To our knowledge, \trainingmethod is the first in LLM personalization to distill both roles separately before POPD, unlike prior privileged distillation methods~\cite{lopezpaz2016GD,chen2019LbC,zhao2026opsd,penaloza2026pidistill,tang2026psd}.

Our contributions are as follows:

\begin{itemize}[leftmargin=1.2em,itemsep=2pt,topsep=2pt]
    \item
We introduce \method, which separates the claims in a dynamic user profile from interpretations of new behavior that are not yet confirmed, holding unresolved interpretations in a Buffer and maintaining per-user commitment rules in a Rulebook.

    \item
We introduce \trainingmethod, which trains a profile manager with POPD after specializing the two roles separately, so that recorded future interactions guide each of the Student's decisions.

    \item
Experiments on three personalization tasks show that \method outperforms multiple baselines under the same Profile budget, that the Profile, the Buffer, and the Rulebook perform best together overall, and that Dual Specialization outperforms one distilled model for both roles and POPD with an undistilled role.

\end{itemize}

%% file: sections/20relatedwork.tex
\begin{figure*}[!t]
  \centering
  \includegraphics[width=\textwidth]{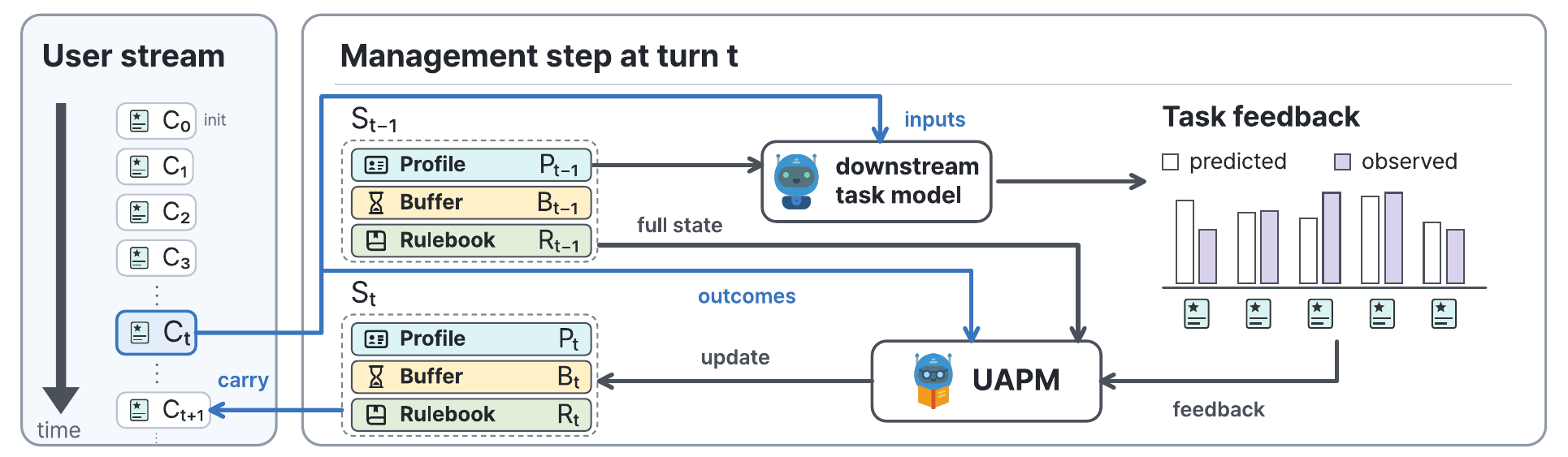}
  \caption{%
\method on a user stream.
Left: chunks arrive in time order.
Right: one management step at turn $t$.
The downstream task model predicts the responses in $C_t$ from the previous Profile $P_{t-1}$.
\method reads the previous state, observed outcomes, and task feedback to update the Profile, Buffer, and Rulebook.
The updated Profile $P_t$ is used on the next chunk $C_{t+1}$.
}
  \Description{Management loop with the Profile, Buffer, and Rulebook, where only the Profile reaches the downstream task model.}
  \label{fig:framework}
\end{figure*}

\subsection{User Profiling}
\label{sec:related-work-user-profiling}
LLM-based user profiling writes a natural-language profile from a user's history.
Static methods construct it once: EXP3RT~\cite{kim2025exp3rt} aggregates review-derived preferences, RLPF~\cite{wu2025rlpf} rewards a summary through next-activity prediction, and LettinGo~\cite{wangl2025lettingo} aligns profile generation with DPO~\cite{rafailov2023dpo} on task performance.
Dynamic methods update it as interactions arrive: PURE~\cite{bang2025pure} appends preferences extracted from each new review, TRIPLE~\cite{noh2025triple} selects among candidate refinements by recent prediction error, DEEPER~\cite{chen-etal-2025-deeper} learns refinement from prediction-error changes across nearby windows, RLPA~\cite{zhao2025rlpa} rewards agreement with simulated users' ground-truth profiles, and RECAP~\cite{zhao2026recap} trains streaming profiles for short-video recommendation with GRPO and adds an interest only when two items support it.
These methods commit each interpretation when it is made and learn, if at all, from later feedback.
\method instead keeps unresolved interpretations in a Buffer outside the Profile and maintains per-user commitment rules in a Rulebook, and it uses the two together to avoid premature commitment.

\subsection{Memory for LLM Agents}
\label{sec:related-work-memory}
LLM agents maintain long-term memory to carry information across long-horizon tasks and sessions.
MemGPT~\cite{packer2023memgpt} revises memory through function calls, MemoryBank~\cite{zhong2024memorybank} adds a forgetting mechanism, Mem0~\cite{chhikara2025mem0} and A-MEM~\cite{xu2025amem} apply add, update, and delete operations or note evolution, and Generative Agents~\cite{park2023generativeagents} and Reflexion~\cite{shinn2023reflexion} store reflections.
ACE~\cite{zhang2026ace} evolves a playbook of task strategies from execution feedback, whereas the Rulebook of \method keeps rules for when to commit, defer, or revise interpretations of one user's behavior and never enters the downstream input.
MEM1~\cite{zhou2026mem1} and MemAgent~\cite{yu2026memagent} use reinforcement learning to iteratively update compact memory states, while Mem-$\alpha$~\cite{wang2025mem-alpha} and Memory-R1~\cite{yan-etal-2026-memoryr1} learn memory operations from downstream outcomes.
Persona\-Mem-v2~\cite{jiang2025personamem-v2} trains a compact per-user memory with reinforcement learning, and BeliefMem~\cite{liao2026beliefmem} exposes probability-weighted candidate beliefs at retrieval.
Unlike BeliefMem, \method keeps unresolved interpretations in a Buffer that the downstream task model never reads, and it trains the manager with future-informed guidance to keep per-user commitment rules.

\subsection{Privileged On-Policy Distillation}
\label{sec:related-work-privileged-on-policy-distillation}
On-policy distillation (OPD)~\cite{agarwal2024gkd} trains a Student on its own outputs with token-level guidance from a Teacher's next-token distributions.
Unlike GRPO~\cite{shao2024grpo}, which samples multiple rollouts and propagates one outcome reward to every step, OPD is cheaper and does not give the same credit to every decision.
Given these advantages, POPD extends OPD so that the Teacher also receives privileged context~\cite{Vladimir2015lupi} that the Student lacks at deployment~\cite{zhao2026opsd,penaloza2026pidistill}.
$\pi$-Distill~\cite{penaloza2026pidistill} and PSD~\cite{tang2026psd} share parameters between the two roles, with PSD using future interactions as privileged context for sequential recommendation, whereas OPSD~\cite{zhao2026opsd} fixes the Teacher at the initial model.
However, using one model for both roles assumes that it can make effective decisions from current information and also use privileged context to guide the Student.
\trainingmethod instead specializes the two roles separately, so each model can focus on the capabilities its role requires, which outperforms one model for both roles (Table~\ref{tab:shared}).

%% file: sections/30methodology.tex
\begingroup
\setlength{\parskip}{0pt}
\setlength{\abovedisplayskip}{4pt plus 1pt minus 1pt}
\setlength{\belowdisplayskip}{4pt plus 1pt minus 1pt}
\setlength{\abovedisplayshortskip}{3pt plus 1pt minus 1pt}
\setlength{\belowdisplayshortskip}{3pt plus 1pt minus 1pt}
\makeatletter
\patchcmd{\paragraph}
  {-.5\baselineskip \@plus -2\p@ \@minus -.2\p@}
  {\z@}
  {}{\PackageError{paper}{Paragraph spacing patch failed}{Check the acmart paragraph definition.}}
\makeatother

In this section, we define the streaming personalization task (Section~\ref{sec:methodology-task-formulation}), present \method for profile management (Section~\ref{sec:methodology-uncertainty-aware-adaptive-profile-manager}), and describe how we train it with \trainingmethod (Section~\ref{sec:methodology-privileged-on-policy-distillation-with-dual-specialization}).

\begin{figure*}[!t]
  \centering
  \includegraphics[width=\textwidth]{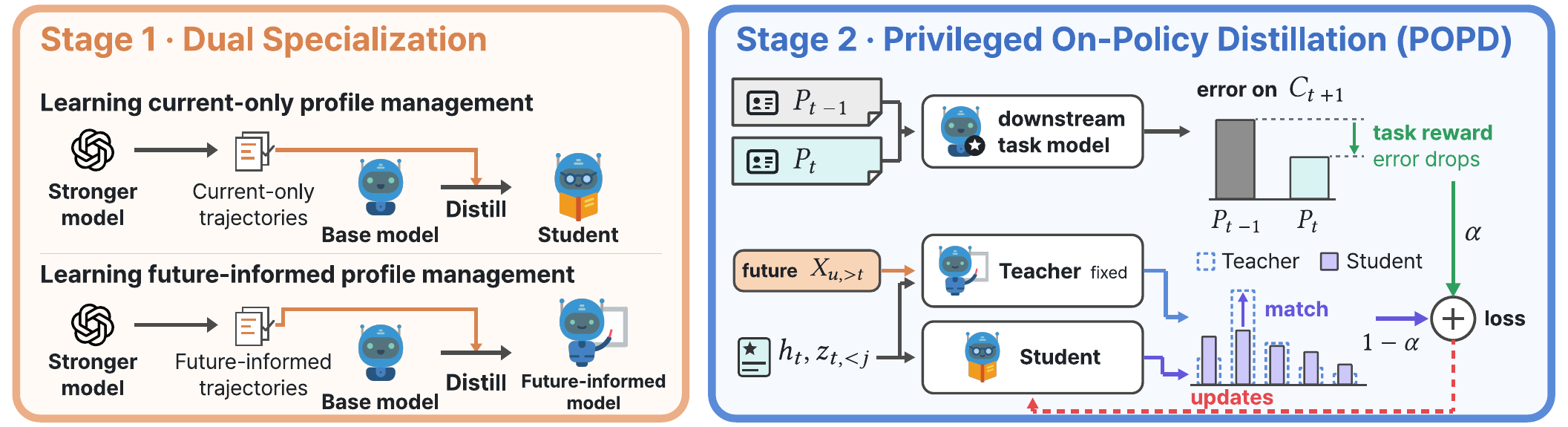}
  \caption{%
Dual Specialization followed by Privileged On-Policy Distillation (POPD).
Stage 1 distills a stronger model's trajectories without and with future interactions into two models, a current-only Student and a future-informed model, respectively.
Stage 2 updates the Student with the task reward and guidance from the future-informed model as the fixed Teacher.
}
  \Description{Two stages: specialization of a current-only Student and a future-informed model, followed by Student updates using the future-informed model as a fixed Teacher, token-distribution matching, and task reward.}
  \label{fig:training}
\end{figure*}

\subsection{Task Formulation}
\label{sec:methodology-task-formulation}
For each user $u$, a chronological stream $X_u=(x_1,x_2,\ldots)$ consists of interactions $x_i=(q_i,a_i)$, each pairing a query with its observed response.
Following prior chunk-based streaming formulations~\cite{chen-etal-2025-deeper}, we process consecutive interactions in \emph{chunks} $C_0,C_1,\ldots$ to reduce sensitivity to noisy feedback, with chunk $C_t$ processed at \emph{turn} $t$.
A profile-management system $\mathcal M$ maintains a user state $S_t$ that includes a compact Profile $P_t$ of the user's characteristics and preferences under a fixed token budget.
The manager writes the initial state $S_0$, including $P_0$, from the initial chunk $C_0$.
At turn $t\geq1$, a downstream task model $f$, trained separately and kept fixed, predicts responses from the previous Profile and the current queries:
{%
\setlength{\abovedisplayskip}{0pt}%
\setlength{\abovedisplayshortskip}{0pt}%
\begin{equation}
  \hat a_i=f(P_{t-1},q_i), \qquad x_i\in C_t.
  \label{eq:predict}
\end{equation}
}
After the responses are observed, $\mathcal M$ updates the state from the previous state, the current chunk, and the predictions $\hat{\mathbf a}_t$:
\begin{equation}
  S_t=\mathcal M(S_{t-1},C_t,\hat{\mathbf a}_t).
  \label{eq:state-update}
\end{equation}
The updated Profile $P_t$ is passed to $f$ at the next turn.
The goal is to maintain a compact Profile that maximizes downstream task performance over the full stream.
An interpretation of new behavior is \emph{unresolved} if the evidence so far supports more than one Profile update and none has been committed yet.

\subsection{\methodfull}
\label{sec:methodology-uncertainty-aware-adaptive-profile-manager}
\method implements $\mathcal M$ with the user state $S_t=(P_t,B_t,R_t)$, consisting of a Profile, a Buffer, and a Rulebook (Figure~\ref{fig:framework}).
\method is uncertainty-aware in that it can mark an interpretation as unresolved by keeping it in the Buffer, outside the Profile, until later interactions separate the readings.
It does not estimate how likely each interpretation is, but decides how the Profile is managed while interpretations remain open.
At each turn, with its model weights fixed, \method generates actions from the previous state, the current interactions, and the task predictions and feedback.
Only the updated Profile reaches the downstream task model. The Buffer and Rulebook are carried forward for later management.

\paragraph{Profile}
The Profile holds user characteristics judged sufficiently supported for downstream use.
Each entry records its scope: \textit{situational} when it holds under a named condition, or \textit{general} when it holds across two or more different conditions.

\paragraph{Buffer}
The Buffer retains hypotheses that are not yet sufficiently supported for the Profile, together with supporting or conflicting evidence and what would settle them.
It targets Uncertain New Behavior: keeping an unresolved interpretation outside the Profile is meant to prevent a premature commitment while later profile management decisions can still use it.

\paragraph{Rulebook}
The Rulebook records user-specific rules for when to commit, defer, or revise an entry, such as how much evidence an interpretation needs before entering the Profile or when conflicting behavior should trigger reconsideration of an entry's scope, along with reflections on earlier feedback that give context for these criteria.
The Rulebook targets user-dependent commitment criteria: because the amount and type of evidence that justifies an update differ across users, the manager writes these criteria per user instead of fixing them in the prompt.
Removing either component lowers performance on almost every metric (Table~\ref{tab:state}).

\paragraph{Actions}
The operations are \textit{add}, \textit{edit}, and \textit{delete} in all three components, and \textit{promote} from the Buffer to the Profile.
An add gives the destination component and the new entry text, an edit gives an existing entry and its replacement text, a delete gives the entry to remove, and a promote gives a Buffer entry to remove and the Profile entry that replaces it.
Profile additions, edits, and promotions also specify general or situational scope, any applicable conditions, and references to supporting evidence.
The management policy $\pi_\theta$ generates an action list $\mathbf d_t$, and applying it updates the user state:
\begin{equation}
  \mathbf d_t\sim\pi_\theta(\cdot\mid S_{t-1},C_t,\hat{\mathbf a}_t),\quad S_t=\mathrm{Apply}(S_{t-1},\mathbf d_t).
  \label{eq:apply}
\end{equation}
Actions are applied in order, skipping any action that exceeds its component budget, refers to a missing entry, or duplicates existing text.
An empty action list, called \textit{keep}, leaves the state unchanged.

\subsection{\trainingmethodfull}
\label{sec:methodology-privileged-on-policy-distillation-with-dual-specialization}

\subsubsection{Dual Specialization}
\label{sec:methodology-dual-specialization}
POPD requires a Student that makes effective management decisions from current and past information and a future-informed model that uses privileged future information to guide those decisions (Figure~\ref{fig:training}).
Shared-backbone designs, common in existing privileged distillation~\cite{penaloza2026pidistill,zhao2026opsd,tang2026psd}, require one model to perform both roles well, but variants using one model for both roles gave weaker task results in our experiments (Table~\ref{tab:shared}).
Dual Specialization therefore trains the two models separately through knowledge distillation~\cite{hinton2015distilling}, each for its own input and role.

A stronger model generates the training targets for both models and receives the remaining future interactions only for future-informed targets.
Eligible targets improve the next-turn task reward over \textit{keep}, or tie with it while changing only the Buffer or Rulebook, and states without an eligible update are discarded.
Ties are retained so that deferrals to the Buffer and Rulebook revisions remain in the targets.
\textit{Keep} itself is never used as a target, because including it made the future-informed model leave the Profile unchanged far more often and lowered the final performance (Appendix~\ref{app:implementation-details-training}).
For a future-informed target, the stronger model sees all remaining interactions while deciding how to update the current state from the current interactions, and its proposals pass the same next-turn eligibility check as those for the Student. The future thus decides which updates are proposed, while the next-turn reward only checks them.
The stronger model also receives a \emph{future-mention instruction} not to write facts that appear only in the remaining interactions.
The future-informed model thus learns to use the future to choose, from the current evidence, updates that also help at the next turn, whereas the Student learns to decide from current information alone.
We serialize each selected action list as tokens $z$ and train each model $\rho\in\{\mathrm S,\mathrm T\}$ by next-token prediction:
{
\begin{equation}
  \mathcal L_{\mathrm{DS}}^{(\rho)}
  =-\frac{1}{N_\rho}\sum_{(h,z)\in\mathcal T_\rho}\sum_{j=1}^{|z|}
  \log\pi_{\theta_\rho}(z_j\mid h,z_{<j}),
  \label{eq:dual-specialization}
\end{equation}
}
where $h$ is the role-specific input, $\mathcal T_\rho$ is its set of selected training targets, and $N_\rho$ counts target tokens. Input tokens are excluded from the loss.
The distilled Student (\studentdistill) initializes $\pi_\theta$ for POPD, and the distilled future-informed model (\teacherdistill) becomes the fixed Teacher $\pi_{\mathrm T}$.

\subsubsection{Privileged On-Policy Distillation (POPD)}
\label{sec:methodology-privileged-on-policy-distillation-popd}
POPD combines task reward with privileged Teacher guidance on the Student's own management decisions.
The task reward measures a step's effect on next-turn predictions relative to \textit{keep}.
For the action list $\mathbf d_t$, let $P_t(\mathbf d_t)$ denote the Profile after applying it to $S_{t-1}$, and let $J(P;C)$ be the task loss on chunk $C$ with Profile $P$:
{
\begin{equation}
  r_t(\mathbf d_t)=J(P_{t-1};C_{t+1})-J(P_t(\mathbf d_t);C_{t+1}).
  \label{eq:reward}
\end{equation}
}
Next-turn reward cannot tell whether an update stays useful later in the stream, so the fixed Teacher guides the decision with the user's remaining recorded interactions and outcomes.
Because future context can reveal facts not yet available at decision time, the Teacher receives the future-mention instruction (Section~\ref{sec:methodology-dual-specialization}).

Both models share the current management input $h_t$, which consists of $S_{t-1}$, $C_t$, and $\hat{\mathbf a}_t$. Task feedback is derived from the observed and predicted responses.
Let $z_t$ denote the token sequence of the action list $\mathbf d_t$.
As the Student outputs a management update, the Teacher receives the same input and Student-generated prefix $z_{t,<j}$, plus the remaining future interactions, and guides the next-token distribution through reverse Kullback--Leibler (KL) divergence~\cite{agarwal2024gkd}:
{
\begin{equation}
  \ell^{\mathrm{KL}}_t=
  \sum_{j\in\mathcal I_t}
  \mathrm{KL}\left(
    \pi_\theta(\cdot\mid h_t,z_{t,<j})
    \,\middle\|\,
    \pi_{\mathrm T}(\cdot\mid h_t,z_{t,<j},X_{u,>t})
  \right),
  \label{eq:kl}
\end{equation}
}
where $\mathcal I_t$ contains eligible decision-token positions, $X_{u,>t}$ contains the interactions in all chunks after turn $t$, and reverse KL weights the guidance by the Student's token distribution.
These positions include the choice of component, so the Teacher can guide whether an interpretation is committed to the Profile or deferred to the Buffer (Appendix~\ref{app:implementation-details-training}).
We use $X_{u,>t}$ in full without truncation, since every Teacher input fits within the model's context length.
Reverse KL is chosen because it performs best overall against forward KL and the Jensen--Shannon divergence (Table~\ref{tab:divergence}).
The objective combines the task loss and $\mathcal L_{\mathrm{KL}}$:
{
\begin{equation}
  \mathcal L_{\mathrm{POPD}}=\alpha\mathcal L_{\mathrm{task}}+(1-\alpha)\mathcal L_{\mathrm{KL}}.
  \label{eq:mix}
\end{equation}
}
Each state samples one action list, whose reward $R_t$ adds skip penalties to $r_t$ and is normalized over the training batch $\mathcal B$:
{
\begin{equation}
  R_t=
  \begin{cases}
    -1 & \text{if }\mathbf d_t\text{ is invalid},\\
    r_t(\mathbf d_t)-n_{\mathrm{skip}}/n_{\mathrm{act}} & \text{otherwise},
  \end{cases}
  \qquad
  A_t=\frac{R_t-\mu_{\mathcal B}}{\sigma_{\mathcal B}},
  \label{eq:advantage}
\end{equation}
}
where $n_{\mathrm{skip}}$ of the $n_{\mathrm{act}}$ actions in $\mathbf d_t$ are not applied (the penalty is 0 for \textit{keep}), and $\mu_{\mathcal B}$ and $\sigma_{\mathcal B}$ are the mean and standard deviation of $R_t$ over $\mathcal B$.
Because each state has one sample, the advantage compares an action list with the other lists in the batch rather than with other samples of the same state~\cite{hu2025reinforce++}, so a list above the batch average receives a positive advantage and one below it a negative advantage.
Because only the Profile reaches $f$, an action list that changes only the Buffer or Rulebook has $r_t=0$. Its advantage then only ranks it against the batch, and guidance on its content comes from $\mathcal L_{\mathrm{KL}}$.
Each action list is used for a single update, and the task term is a policy-gradient loss over its tokens weighted by $A_t$:
{
\begin{equation}
  \mathcal L_{\mathrm{task}}=-\frac{1}{|\mathcal B|}\sum_{t\in\mathcal B}\frac{A_t}{D}\sum_{j=1}^{|z_t|}w_{t,j}\log\pi_\theta(z_{t,j}\mid h_t,z_{t,<j}),
  \label{eq:task}
\end{equation}
}
where $\mathcal B$ is the set of trainable action lists in the update batch and $D$ is a fixed token count (Appendix~\ref{app:implementation-details-training}).
The weight $w_{t,j}=\min\{3,\pi_\theta(z_{t,j})/\pi_{\mathrm{gen}}(z_{t,j})\}$, with the conditioning on $h_t$ and $z_{t,<j}$ omitted, is held constant in the gradient.
Here $\pi_{\mathrm{gen}}$ is the probability that the generation engine assigned when sampling $z_t$ with the same weights as $\pi_\theta$.
In Eq.~\eqref{eq:mix}, $\mathcal L_{\mathrm{KL}}$ averages $\ell^{\mathrm{KL}}_t$ from Eq.~\eqref{eq:kl}, divided by $D$, over the action lists in $\mathcal B$.
The loss raises the probability of every token of an action list with a positive advantage and lowers it for one with a negative advantage.
The token sum is divided by the fixed $D$ rather than by $|z_t|$, so the update per token does not shrink for longer lists~\cite{liu2025drgrpo}.
The weight $w_{t,j}$ corrects for the generation engine assigning slightly different probabilities to the sampled tokens than the trainer, and its cap bounds this correction~\cite{yao2025tis}.
The mixing weight $\alpha$ balances task feedback and Teacher guidance (Appendix~\ref{app:implementation-details-training}), and Appendix~\ref{app:implementation-details-task-reward} gives the task-specific rewards.

\endgroup

%% file: sections/40experiments.tex
The experiments examine how much \method improves personalization under the same Profile budget (Section~\ref{sec:experiments-main-results}), how much its training stages and state components contribute (Sections~\ref{sec:experiments-training-stage-ablation} to~\ref{sec:experiments-popd-analysis}), and how it handles uncertain signals (Sections~\ref{sec:experiments-buffer-rulebook-usage} and~\ref{sec:experiments-uncertainty-handling-analysis}).

\subsection{Experimental Setup}
\label{sec:experiments-experimental-setup}

\subsubsection{Datasets}
\label{sec:experiments-datasets}
\setcounter{topnumber}{1}
\suppressfloats[t]
\input{tables/dataset_statistics}
We evaluate \method on three personalization tasks, each organized as a time-ordered user stream: rating prediction on Amazon Books~\cite{hou2024amazon} (ratings 1 to 5, 105 interactions per user), movie tagging on LaMP-2~\cite{salemi2023lamp} (15 tags, 10 to 105 interactions per user), and answer generation from titles on LaMP-QA~\cite{salemi-zamani-2025-lamp}, extended to 25 interactions per user (Appendix~\ref{app:datasets}).
\method splits its training users into disjoint subsets for Dual Specialization, POPD, and the downstream task model, and trained baselines use the first two (Table~\ref{tab:dataset}).
Each dataset is divided by user into training, validation, and test sets.
After five initialization interactions, we manage the state in chunks of five on Amazon Books and of one on LaMP-2 and LaMP-QA, and evaluate on the remaining turns of each stream.

\subsubsection{Baselines}
\label{sec:experiments-baselines}
The baselines cover history inputs (No Personalization and Recent History), static profiling (Static Profile Prompting, RLPF~\cite{wu2025rlpf}, and LettinGo~\cite{wangl2025lettingo}), dynamic profiling (Dynamic Profile Prompting, PURE~\cite{bang2025pure}, TRIPLE~\cite{noh2025triple}, and DEEPER~\cite{chen-etal-2025-deeper}), and memory management (MemGPT~\cite{packer2023memgpt}, ACE~\cite{zhang2026ace}, and MemAgent~\cite{yu2026memagent}, whose fixed-size memory serves as the Profile), with Full History as a reference outside the budget (Appendix~\ref{app:baselines}).
All baselines use the same backbone as \method and, except Full History, the same downstream input budget, and every method, including the history inputs, has its own fine-tuned downstream task model, as for \method.
Only the manager reads the Buffer and the Rulebook, so they raise its cost (Appendix~\ref{app:implementation-details-compute}) but not the downstream input, and Section~\ref{sec:experiments-state-component-ablation} tests whether their gain comes from capacity alone.
\subsubsection{Metrics}
\label{sec:experiments-metrics}
Amazon Books uses MAE and RMSE, and LaMP-2 uses accuracy (Acc) and Macro-F1.
LaMP-QA follows the benchmark's aspect-based evaluation~\cite{salemi-zamani-2025-lamp}: an LLM judge, Claude Haiku 5.5\footnote{https://www.anthropic.com/claude-haiku-5-5} with thinking disabled, rates from 0 to 2 how well an answer addresses each aspect, a need stated in the question body, and the score is the mean rating divided by 2 (one minus it is $J$ in training).
To check this judge, GPT-5.6-terra\footnote{https://openai.com/index/gpt-5-6/}, which generates no training targets, re-judges 200 sampled test questions, and the two judges rank the methods similarly on these questions (Kendall's $\tau_b=0.73$, Spearman's $\rho=0.90$, Appendix~\ref{app:implementation-details-judge}).

\subsubsection{Implementation Details}
\label{sec:experiments-implementation-details}
We use Qwen3.5-4B\footnote{https://qwen.ai/blog?id=qwen3.5} with reasoning disabled as the backbone of the profile manager and the downstream task model, and GPT-6.1-sol\footnote{https://openai.com/index/introducing-gpt-6-1-sol/} with low reasoning effort to generate the supervised targets.
The Profile budget is 128 tokens on Amazon Books and LaMP-2 and 256 tokens on LaMP-QA, where the larger budget accommodates the longer inputs of long-form question answering. Recent History receives the same budget, and the Buffer and the Rulebook receive 1,024 tokens each.
POPD uses $\alpha=0.5$, and the downstream task model is fine-tuned separately for each method, with separate downstream task models for the task reward and the final evaluation.
Management outputs are sampled at temperature 1 in training and testing.
For every trained model, test results use the checkpoint that performs best on the validation users.
Appendix~\ref{app:implementation-details} gives training details, hyperparameters, and task rewards. All prompts are released with the code.

\input{tables/main_results}
\input{tables/training_stages}
\subsection{Main Results}
\label{sec:experiments-main-results}
\textbf{\method improves personalization the most overall under the same Profile budget} (Table~\ref{tab:main})\textbf{.}
Its margin over the strongest same-budget baseline is modest but consistent.
Together with the ablations (Section~\ref{sec:experiments-state-component-ablation}), this suggests that keeping unresolved interpretations out of the Profile and keeping per-user criteria for committing them make Profile updates more effective.
Full History, a reference outside the budget, remains stronger on Amazon Books and LaMP-QA but not on LaMP-2, where a refined Profile helps more.
However, it costs about 9 to 24 times the Profile budget in tokens, grows with the stream, and is hard to organize and maintain~\cite{ju2026bump}.

\textbf{Updating the Profile helps only when the updates are made well.}
Dynamic profiling does not consistently outperform static profiling: most dynamic baselines outperform static ones on Amazon Books, but on LaMP-2 and LaMP-QA every static method outperforms every dynamic baseline.
Updating can help, but poorly managed updates can leave the Profile less useful than one that is never updated.
Training does not always help either: DEEPER is below Dynamic Profile Prompting on every metric, and LettinGo is below Static Profile Prompting in MAE on Amazon Books.
These methods are trained on far less data than in their original settings~(Appendix~\ref{app:implementation-details-adapted-baselines}), which may limit what their training objectives can learn.
Memory management does not clearly outperform profiling either: MemGPT improves on Dynamic Profile Prompting only on LaMP-QA while reading about 20,000 input tokens per turn (Appendix~\ref{app:implementation-details-compute}), and ACE and MemAgent remain below every static method on LaMP-2.
One possible reason is that these methods target multi-session conversation~\cite{packer2023memgpt}, agent reasoning~\cite{zhang2026ace}, or long-document question answering~\cite{yu2026memagent}, where what later steps need is relatively clear, whereas in personalization it is uncertain which interactions will matter for later predictions.
The datasets also differ: Recent History outperforms most profiling methods on Amazon Books but not on LaMP-2, and on LaMP-QA most personalized inputs score even below No Personalization, as in the original benchmark~\cite{salemi-zamani-2025-lamp}, suggesting that most methods do not capture the personalization signal of long-form community answers.

\subsection{Training Stage and Role Specialization Ablation}
\label{sec:experiments-training-stage-ablation}
To examine the contribution of the training stages and of role specialization, the distillation of each role is removed, or one shared model replaces the two specialized roles (Table~\ref{tab:signal}).
Prompting uses the untrained backbone, and POPD rows pair a prompted or distilled Student with a prompted or distilled Teacher.

\textbf{Training is needed for the manager to use its state and information.}
With the full state but no training, Student Prompting is worse than Dynamic Profile Prompting (Table~\ref{tab:main}) on LaMP-2 and in MAE on Amazon Books, so without training the full state does not consistently help.
\studentdistill outperforms Student Prompting on every metric, so distillation teaches the Student to revise the current state.
With the same future input, the distilled future-informed model also outperforms the prompted one, so distillation teaches it to use the future when revising the current state.
It outperforms \method on every metric and serves as a reference point for a manager that observes the future, which \method approaches on LaMP-2 and LaMP-QA.
\studentdistill alone matches the Stage 1 target generator (GPT-6.1-sol with \method's prompt and state) on Amazon Books but not on LaMP-2 and LaMP-QA, whereas POPD with both distilled roles outperforms it on every metric.

\textbf{POPD performs best with both roles distilled, and the Student's distillation matters more.}
With both roles distilled, POPD outperforms \studentdistill and every POPD setting with an undistilled role on all three datasets.
Removing \studentdistill lowers performance more than removing \teacherdistill on every metric, so the Student's specialization contributes more, although both are needed for the best result.
Without either distillation (OPSD from the backbone), POPD performs worst among the POPD settings on every metric.

\textbf{One model for both roles performs worse than Dual Specialization.}
OPSD~\cite{zhao2026opsd} initializes the Student and a fixed Teacher from the same model and gives only the Teacher privileged context.
Two shared-backbone variants follow OPSD with \studentdistill or \teacherdistill, and multi-task distillation trains one model on both roles' targets and uses the current Student with future input as the Teacher, as in $\pi$-Distill~\cite{penaloza2026pidistill}.
\method outperforms these variants on all three datasets, which supports specializing the Student and the Teacher separately.

\input{tables/state_ablation}
\input{tables/popd_analysis}
\subsection{State Component Ablation}
\label{sec:experiments-state-component-ablation}
\textbf{The full state performs best overall, and the Buffer and the Rulebook are most useful together.}
To examine the contribution of each state component, the Buffer, the Rulebook, or both are removed from the prompt, targets, and state, and both training stages are rerun (Table~\ref{tab:state}).
Without the Buffer, unresolved interpretations have no place to wait, so the manager must commit them to the Profile or drop them.
Every variant goes through the same distillation and POPD as \method, so \method's gain over the Profile-only variant (\method w/o Buffer and Rulebook) is not due to distillation alone.
In an additional experiment on LaMP-2, replacing the Buffer and the Rulebook with one unstructured notepad of the same 2,048 tokens (Appendix~\ref{app:additional-analyses}), trained with the same stronger model and stages, gives 0.5040 accuracy and 0.4456 Macro-F1, recovering about half of the gain of the full state over the Profile-only variant (Table~\ref{tab:state}), so capacity alone does not explain the gain on LaMP-2.
Measured by the gain over removing both, the Rulebook contributes more than the Buffer on Amazon Books and the Buffer contributes more on LaMP-QA, whereas the two are close on LaMP-2.
Either component alone, however, gives only marginal gains: on LaMP-2 and LaMP-QA it recovers less than half of the gain of the full state, and on Amazon Books the Buffer alone gives a higher MAE than removing both.
One possible interpretation is that the Buffer alone lacks user-specific criteria for its hypotheses and the Rulebook alone lacks a place to keep ambiguous interpretations outside the Profile, so the two work best together.
The exception is RMSE on Amazon Books, where removing the Buffer, like Teacher guidance alone ($\alpha=0$ in Table~\ref{tab:alpha}), gives a lower RMSE but a higher MAE, possibly because committing interpretations to the Profile instead of deferring them avoids some large errors but adds small ones.

\begin{table}[t]\begingroup\renewenvironment{table}[1][]{}{}\input{tables/component_usage}\par\vspace{10pt}\input{tables/ambiguous_events}\endgroup\end{table}
\subsection{POPD Analysis}
\label{sec:experiments-popd-analysis}
\label{sec:experiments-kl-divergence}
\textbf{Reverse KL performs best overall, although the differences among the divergences are not large.}
To examine the design of POPD, the divergence, the mixing weight, and the Teacher's privileged input are varied (Table~\ref{tab:popd}).
Forward KL makes the Student cover every decision the Teacher considers likely, and JSD balances the two directions. One reason may be that reverse KL penalizes only Student decisions the Teacher finds unlikely~\cite{agarwal2024gkd}, and since the Teacher relies on a future the Student cannot see, pruning its poor decisions suits it better than covering every Teacher decision.

\label{sec:experiments-mixing-weight}
\textbf{The equal weight $\alpha=0.5$ performs best overall in our runs.}
The weight $\alpha$ in Eq.~\eqref{eq:mix} balances next-turn task reward ($\alpha=1$) against future-informed Teacher guidance ($\alpha=0$).
Either signal alone is worse than $\alpha=0.5$ on most metrics, suggesting that next-turn task reward and Teacher guidance complement each other, although small gaps may reflect training variance (Appendix~\ref{sec:limitations}).

\label{sec:experiments-privileged-information}
\textbf{Removing either the future interactions from the Teacher's input or the future-mention instruction from its prompt lowers performance on every metric.}
Without future interactions, the Teacher sees only the Student's input, which turns POPD into on-policy distillation~\cite{agarwal2024gkd} from a distilled Teacher.
Because the Teacher is specialized with future input, removing that input also shifts its input distribution, but in an additional experiment, a Teacher specialized without future input from the start, which avoids this shift, still gives lower accuracy and Macro-F1 on LaMP-2 (0.5170 and 0.4675) than \method (0.5313 and 0.4742), so the input shift alone does not explain the gain from future input.
A Teacher that receives the future but is not specialized to use it, as in \studentdistill for both roles (Table~\ref{tab:shared}), also performs worse than \method on all three datasets, suggesting that the gain needs both the future input and a Teacher trained to use it.
Removing the future-mention instruction also lowers performance, although in neither setting does the Student name a future item verbatim on Amazon Books or LaMP-QA.
Without the instruction, the Teacher may favor decisions only the future supports, which the Student cannot reproduce, though removal changes its prompt.

\subsection{Buffer and Rulebook Usage}
\label{sec:experiments-buffer-rulebook-usage}
\textbf{The Buffer holds ambiguous evidence outside the Profile, and replacing a user's Rulebook with another user's lowers performance.}
To examine how the manager uses its state, the applied operations on each component are counted over all test streams (Table~\ref{tab:usage}).
Most operations add or edit entries, and deletions are rare, especially in the Profile, so the manager mostly refines entries instead of deleting them.
On Amazon Books and LaMP-2, more new entries go to the Buffer as hypotheses than to the Profile, whereas on LaMP-QA more go to the Profile directly, likely because a question states much of what its answer should address while ratings and tags leave preferences implicit.
Because the downstream task model reads only the Profile, these hypotheses do not enter the downstream input while they remain unresolved.
Hypotheses are rarely promoted as they are, and Profile actions cite Buffer entries far more often, so held hypotheses mainly serve as evidence for Profile actions.
Most hypotheses are still held at stream end.

Replacing a user's Rulebook at every turn with a random test user's lowers all five metrics, consistent with rules written for each user's own situations, although part of the drop may come from a mismatch with the user's Profile and Buffer.
In addition, different users' Rulebooks are less similar to each other than one user's Rulebooks over time (Appendix~\ref{app:additional-analyses}).

\subsection{Performance after Ambiguous Evidence}
\label{sec:experiments-uncertainty-handling-analysis}
\textbf{After ambiguous evidence, \method performs best on all five metrics against the strongest same-budget baselines and its own ablations.}
To examine ambiguous evidence, predictions over the five interactions starting from the first management turn after a low rating in a category the user usually rates highly, a new tag, or a question from a new community are compared (Table~\ref{tab:uncertainty}, Appendix~\ref{app:ambiguous-events}).
\method performs best after these events, whereas Static Profile Prompting, whose Profile never absorbs new evidence, performs worst on Amazon Books, and TRIPLE performs worst on LaMP-2 and LaMP-QA.
Removing the Buffer or the Rulebook lowers performance after these events on every metric. On Amazon Books and LaMP-QA, this drop is about twice that over all predictions (Table~\ref{tab:state}) or more, whereas on LaMP-2 it is not larger. Without the Buffer, Amazon Books RMSE is lower overall (Table~\ref{tab:state}) but higher after ambiguous events.
On Amazon Books and LaMP-QA, each component thus matters more after ambiguous evidence than overall.

%% file: tables/dataset_statistics.tex
\begin{table}[t]
\centering
\footnotesize
\setlength{\tabcolsep}{2.6pt}
\caption{Dataset statistics. DS, POPD, and DTM are the disjoint training subsets for Dual Specialization, POPD, and the downstream task model; $L_P$ is the Profile budget in tokens.}
\vspace{-2pt}
\begin{tabular}{l l l ccccc c}
  \toprule
  \textbf{Dataset} & \textbf{Task} & \textbf{Unit} & \textbf{DS} & \textbf{POPD} & \textbf{DTM} & \textbf{Val} & \textbf{Test} & $L_P$ \\
  \midrule
  \multirow{2}{*}{Amazon Books} & \multirow{2}{*}{rating} & users & 400 & 400 & 133 & 50 & 50 & \multirow{2}{*}{128} \\
   & & instances & 42,000 & 42,000 & 13,965 & 5,250 & 5,250 & \\
  \midrule
  \multirow{2}{*}{LaMP-2} & \multirow{2}{*}{tagging} & users & 129 & 129 & 43 & 51 & 77 & \multirow{2}{*}{128} \\
   & & instances & 3,759 & 3,794 & 1,170 & 1,380 & 2,222 & \\
  \midrule
  \multirow{2}{*}{LaMP-QA} & \multirow{2}{*}{generation} & users & 400 & 400 & 133 & 50 & 50 & \multirow{2}{*}{256} \\
   & & instances & 10,000 & 10,000 & 3,325 & 1,250 & 1,250 & \\
  \bottomrule
\end{tabular}
\label{tab:dataset}
\end{table}

%% file: tables/main_results.tex
\providecommand{\sd}[1]{\textsubscript{$\pm$#1}}
\begin{table}[t]
\centering
\footnotesize
\caption{Main results. Methods above the double rule use the same backbone and Profile budget. Values with subscripts are means and standard deviations over three sampling runs. $^\dagger$ denotes a statistically significant improvement over the best baseline above the double rule (paired t-test over test users on run-averaged metrics, $p<0.05$).}
\setlength{\tabcolsep}{1.5pt}
\resizebox{\columnwidth}{!}{%
\begin{tabular}{l cc cc c}
\toprule
\multirow{2.5}{*}{\textbf{Method}} & \multicolumn{2}{c}{\textbf{Amazon Books}} & \multicolumn{2}{c}{\textbf{LaMP-2}} & \textbf{LaMP-QA} \\
\cmidrule(lr){2-3} \cmidrule(lr){4-5} \cmidrule(lr){6-6}
 & MAE$\downarrow$ & RMSE$\downarrow$ & Acc$\uparrow$ & Macro-F1$\uparrow$ & Score$\uparrow$ \\
\midrule
No Personalization & 0.7431\sd{0.000} & 0.9730\sd{0.000} & 0.4614\sd{0.000} & 0.3746\sd{0.000} & 0.3976\sd{0.000} \\
\midrule
Recent History & 0.6213\sd{0.000} & 0.9199\sd{0.000} & 0.4909\sd{0.000} & 0.4202\sd{0.000} & 0.3619\sd{0.000} \\
\midrule
\grouprow{6}{Static user profiling}
Prompting & 0.6629\sd{0.005} & 1.0073\sd{0.005} & 0.5006\sd{0.002} & 0.4429\sd{0.002} & 0.3673\sd{0.002} \\
RLPF & 0.6550\sd{0.005} & 0.9937\sd{0.005} & 0.5017\sd{0.001} & 0.4456\sd{0.002} & \underline{0.4023}\sd{0.002} \\
LettinGo & 0.6724\sd{0.005} & 0.9723\sd{0.005} & \underline{0.5063}\sd{0.002} & \underline{0.4545}\sd{0.002} & 0.3830\sd{0.002} \\
\midrule
\grouprow{6}{Dynamic user profiling}
Prompting & 0.6499\sd{0.005} & 0.9452\sd{0.005} & 0.4932\sd{0.001} & 0.4293\sd{0.001} & 0.3330\sd{0.002} \\
PURE & 0.6965\sd{0.005} & 1.0001\sd{0.005} & 0.4989\sd{0.001} & 0.4353\sd{0.001} & 0.3409\sd{0.002} \\
TRIPLE & \underline{0.5940}\sd{0.005} & \underline{0.9195}\sd{0.005} & 0.4727\sd{0.002} & 0.4301\sd{0.002} & 0.2833\sd{0.002} \\
DEEPER & 0.6513\sd{0.004} & 0.9678\sd{0.005} & 0.4750\sd{0.002} & 0.4275\sd{0.002} & 0.3292\sd{0.002} \\
\midrule
\grouprow{6}{Memory management}
MemGPT & 0.6819\sd{0.006} & 0.9834\sd{0.005} & 0.4795\sd{0.002} & 0.4242\sd{0.002} & 0.3624\sd{0.002} \\
ACE & 0.6801\sd{0.005} & 1.0018\sd{0.005} & 0.4940\sd{0.001} & 0.4419\sd{0.001} & 0.3218\sd{0.002} \\
MemAgent & 0.6065\sd{0.004} & 0.9909\sd{0.004} & 0.4900\sd{0.002} & 0.4401\sd{0.002} & 0.3675\sd{0.003} \\
\midrule
\method & \textbf{0.5801}$^\dagger$\sd{0.005} & \textbf{0.9152}\sd{0.005} & \textbf{0.5313}$^\dagger$\sd{0.002} & \textbf{0.4742}$^\dagger$\sd{0.002} & \textbf{0.4163}$^\dagger$\sd{0.002} \\
\midrule
\midrule
Full History & 0.5712\sd{0.000} & 0.8800\sd{0.000} & 0.5193\sd{0.000} & 0.4566\sd{0.000} & 0.4430\sd{0.000} \\
\bottomrule
\end{tabular}}
\label{tab:main}
\end{table}

%% file: tables/training_stages.tex
\begin{table}[t]
\centering
\footnotesize
\caption{Training stage and role specialization ablation. Future-informed rows see future input and are not ranked.}
\setlength{\tabcolsep}{2.5pt}
\resizebox{\columnwidth}{!}{%
\begin{tabular}{l cc cc c}
\toprule
\multirow{2.5}{*}{\textbf{Method}} & \multicolumn{2}{c}{\textbf{Amazon Books}} & \multicolumn{2}{c}{\textbf{LaMP-2}} & \textbf{LaMP-QA} \\
\cmidrule(lr){2-3} \cmidrule(lr){4-5} \cmidrule(lr){6-6}
 & MAE$\downarrow$ & RMSE$\downarrow$ & Acc$\uparrow$ & Macro-F1$\uparrow$ & Score$\uparrow$ \\
\midrule
\grouprow{6}{Stage 1 target generator with current input (not ranked)}
Stronger Model (GPT-6.1-sol) & 0.5889 & 0.9297 & 0.5205 & 0.4698 & 0.4038 \\
\midrule
\grouprow{6}{Student with current input}
Student Prompting & 0.7075 & 0.9342 & 0.4670 & 0.4142 & 0.3695 \\
\studentdistill & 0.5884 & 0.9236 & 0.5165 & 0.4543 & 0.3919 \\
\midrule
\grouprow{6}{Future-informed model with current and future input (not ranked)}
Future-informed Prompting (+future) & 0.6648 & 0.9721 & 0.4898 & 0.4303 & 0.4021\\
\teacherdistill (+future) & 0.5666 & 0.8924 & 0.5330 & 0.4751 & 0.4181\\
\midrule
\grouprow{6}{POPD Student with current input}
POPD w/o either distillation & 0.7033 & 0.9842 & 0.4830 & 0.4257 & 0.3694 \\
POPD w/o \studentdistill & 0.6849 & 0.9362 & 0.4875 & 0.4269 & 0.3712 \\
POPD w/o \teacherdistill & \underline{0.5853} & 0.9201 & \underline{0.5172} & \underline{0.4580} & 0.3954 \\
\method & \textbf{0.5801} & \textbf{0.9152} & \textbf{0.5313} & \textbf{0.4742} & \textbf{0.4163} \\
\midrule
\grouprow{6}{One distilled model for both roles}
\studentdistill for both roles & 0.5932 & 0.9255 & 0.5153 & 0.4562 & 0.3912\\
\teacherdistill for both roles & 0.5889 & \underline{0.9181} & 0.5108 & 0.4529 & 0.4007\\
Multi-task distillation for both roles & 0.6005 & 0.9182 & 0.5148 & 0.4558 & \underline{0.4038} \\
\bottomrule
\end{tabular}}
\label{tab:signal}\label{tab:shared}
\end{table}

%% file: tables/state_ablation.tex
\begin{table}[t]
\centering
\footnotesize
\caption{State component ablation. Best and second-best results are in bold and underlined.}
\setlength{\tabcolsep}{1.5pt}
\resizebox{\columnwidth}{!}{%
\begin{tabular}{l cc cc c}
\toprule
\multirow{2.5}{*}{\textbf{Condition}} & \multicolumn{2}{c}{\textbf{Amazon Books}} & \multicolumn{2}{c}{\textbf{LaMP-2}} & \textbf{LaMP-QA} \\
\cmidrule(lr){2-3} \cmidrule(lr){4-5} \cmidrule(lr){6-6}
 & MAE$\downarrow$ & RMSE$\downarrow$ & Acc$\uparrow$ & Macro-F1$\uparrow$ & Score$\uparrow$ \\
\midrule
\method & \textbf{0.5801} & \underline{0.9152} & \textbf{0.5313} & \textbf{0.4742} & \textbf{0.4163} \\
\method w/o Buffer and Rulebook & 0.5961 & 0.9432 & 0.4818 & 0.4150 & 0.3834\\
\method w/o Rulebook & 0.6044 & 0.9347 & 0.4852 & 0.4169 & \underline{0.3965} \\
\method w/o Buffer & \underline{0.5904} & \textbf{0.9080} & \underline{0.4864} & \underline{0.4188} & 0.3902 \\
\bottomrule
\end{tabular}}
\label{tab:state}
\end{table}

%% file: tables/popd_analysis.tex
\begin{table}[t]
\centering
\footnotesize
\caption{POPD analysis: divergence, mixing weight, and Teacher input. Best and second-best are marked per group.}
\setlength{\tabcolsep}{1.5pt}
\resizebox{\columnwidth}{!}{%
\begin{tabular}{l cc cc c}
\toprule
\multirow{2.5}{*}{\textbf{Setting}} & \multicolumn{2}{c}{\textbf{Amazon Books}} & \multicolumn{2}{c}{\textbf{LaMP-2}} & \textbf{LaMP-QA} \\
\cmidrule(lr){2-3} \cmidrule(lr){4-5} \cmidrule(lr){6-6}
 & MAE$\downarrow$ & RMSE$\downarrow$ & Acc$\uparrow$ & Macro-F1$\uparrow$ & Score$\uparrow$ \\
\midrule
\grouprow{6}{Divergence}
Forward KL & \underline{0.5915} & \underline{0.9227} & 0.5250 & 0.4627 & \underline{0.3993}\\
JSD & 0.6004 & 0.9357 & \underline{0.5290} & \textbf{0.4755} & 0.3937\\
Reverse KL (\method) & \textbf{0.5801} & \textbf{0.9152} & \textbf{0.5313} & \underline{0.4742} & \textbf{0.4163} \\
\midrule
\grouprow{6}{Mixing weight $\alpha$}
$\alpha=0$ (Teacher guidance only) & 0.6063 & \textbf{0.9084} & 0.5148 & 0.4544 & 0.4060\\
$\alpha=0.25$ & 0.5960 & 0.9233 & 0.5170 & 0.4581 & \underline{0.4070}\\
$\alpha=0.5$ (\method) & \textbf{0.5801} & \underline{0.9152} & \textbf{0.5313} & \textbf{0.4742} & \textbf{0.4163} \\
$\alpha=0.75$ & 0.5929 & 0.9178 & \underline{0.5267} & \underline{0.4673} & 0.3936\\
$\alpha=1$ (task reward only) & \underline{0.5895} & 0.9194 & 0.5210 & 0.4617 & 0.3916\\
\midrule
\grouprow{6}{Teacher input}
\method (future context and instruction) & \textbf{0.5801} & \textbf{0.9152} & \textbf{0.5313} & \textbf{0.4742} & \textbf{0.4163} \\
without future context & 0.5978 & \underline{0.9177} & \underline{0.5244} & \underline{0.4643} & 0.3846\\
without future-mention instruction & \underline{0.5879} & 0.9339 & 0.5148 & 0.4593 & \underline{0.3917}\\
\bottomrule
\end{tabular}}
\label{tab:popd}\label{tab:divergence}\label{tab:alpha}\label{tab:teacher}
\end{table}

%% file: tables/component_usage.tex
\begin{table}[t]
\centering
\footnotesize
\caption{Operations per state component, as total counts of applied operations over all test streams. In the last group, percentages are shares of the Buffer hypotheses added, or of all applied Profile actions in the last row.}
\setlength{\tabcolsep}{4pt}
\resizebox{\columnwidth}{!}{%
\begin{tabular}{l ccc}
\toprule
\textbf{Component} & \textbf{Amazon Books} & \textbf{LaMP-2} & \textbf{LaMP-QA} \\
\midrule
\grouprow{4}{add / edit / delete}
Profile & 153 / 1,122 / 5 & 272 / 1,388 / 21 & 350 / 136 / 4 \\
Buffer & 942 / 363 / 182 & 405 / 247 / 141 & 193 / 180 / 51 \\
Rulebook & 971 / 131 / 29 & 789 / 978 / 185 & 268 / 158 / 49 \\
\midrule
Promote (Buffer to Profile) & 21 & 7 & 11 \\
Management turns & 1,000 & 1,837 & 1,000 \\
Keep (no action applied) & 2 (0.2\%) & 19 (1.0\%) & 13 (1.3\%) \\
\midrule
\grouprow{4}{Buffer hypotheses}
Held at stream end & 739 (78.5\%) & 257 (63.5\%) & 131 (67.9\%) \\
Edited while held & 250 (26.5\%) & 154 (38.0\%) & 67 (34.7\%) \\
Profile actions citing it & 199 (15.3\%) & 303 (18.0\%) & 79 (15.8\%) \\
\bottomrule
\end{tabular}}
\label{tab:usage}
\end{table}

%% file: tables/ambiguous_events.tex
\begin{table}[t]
\centering
\footnotesize
\caption{Performance over the five interactions starting from the first management turn after an ambiguous event.}
\setlength{\tabcolsep}{4pt}
\resizebox{\columnwidth}{!}{%
\begin{tabular}{l cc cc c}
\toprule
\multirow{2.5}{*}{\textbf{Method}} & \multicolumn{2}{c}{\textbf{Amazon Books}} & \multicolumn{2}{c}{\textbf{LaMP-2}} & \textbf{LaMP-QA} \\
\cmidrule(lr){2-3} \cmidrule(lr){4-5} \cmidrule(lr){6-6}
 & MAE$\downarrow$ & RMSE$\downarrow$ & Acc$\uparrow$ & Macro-F1$\uparrow$ & Score$\uparrow$ \\
\midrule
Static Profile Prompting & 0.8447 & 1.1867 & 0.4456 & 0.3981 & 0.3547 \\
RLPF & 0.8006 & 1.1602 & 0.4593 & 0.4129 & \underline{0.4020} \\
LettinGo & 0.7779 & 1.0487 & 0.4548 & \underline{0.4136} & 0.3700 \\
Dynamic Profile Prompting & 0.7931 & 1.1129 & \underline{0.4654} & 0.4093 & 0.3044 \\
TRIPLE & 0.7414 & \underline{0.9995} & 0.4312 & 0.3915 & 0.2929 \\
\midrule
\method w/o Buffer & \underline{0.7194} & 1.0012 & 0.4624 & 0.3985 & 0.3655 \\
\method w/o Rulebook & 0.7305 & 1.0265 & 0.4616 & 0.3988 & 0.3851 \\
\method w/o Buffer and Rulebook & 0.7342 & 1.0323 & 0.4563 & 0.3946 & 0.3887 \\
\method & \textbf{0.6871} & \textbf{0.9832} & \textbf{0.4966} & \textbf{0.4477} & \textbf{0.4231} \\
\bottomrule
\end{tabular}}
\label{tab:uncertainty}
\end{table}

%% file: sections/50conclusion.tex
This paper introduced \method, which separates the claims in a dynamic user profile from unconfirmed interpretations of new behavior through a Profile, a Buffer, and a Rulebook, and \trainingmethod, which trains it by Dual Specialization followed by POPD with a future-informed Teacher.
Results and analyses on three Web personalization datasets show that \method improves personalization under the same Profile budget and performs best after ambiguous evidence among the compared methods.
These results suggest that holding unresolved interpretations outside the Profile can make Web user profiles more useful for personalization, and that streaming personalization can treat unresolved interpretations as part of the managed user state.
Appendices~\ref{sec:limitations} and~\ref{sec:ethical-considerations} discuss the limitations and ethical considerations of this work.

%% file: sections/71limitations.tex
We evaluate offline on public datasets, without a user study.
Dual Specialization keeps only Profile changes that improve the next turn and may discard ones that pay off later, which POPD only partly offsets. A multi-turn target filter is left for future work.
The manager of \method spends more input tokens per management turn than simpler dynamic methods such as Dynamic Profile Prompting, and more output tokens on two of the three datasets, although far fewer than MemGPT (Table~\ref{tab:cost}). This cost grows with the number of users, in exchange for higher performance.
The LaMP-QA judge, which also gives the training reward, is not validated against human ratings in this study.
However, LLM judges of this kind have been validated against human judgments on personalized and long-form answers~\cite{salemi-zamani-2025-lamp,kim2026ipqa}, and the judge ranks the methods similarly to GPT-5.6-terra (Section~\ref{sec:experiments-metrics}).
Baselines match the downstream input of \method, not its manager state with the Buffer and the Rulebook. Given the training cost, the notepad variant (Section~\ref{sec:experiments-state-component-ablation}), which matches this state, and the Teacher specialized without future input (Section~\ref{sec:experiments-privileged-information}) are run only on LaMP-2. Streams have at most 105 interactions, and longer streams may need a way to remove Buffer hypotheses that stay unresolved for long.
The three runs measure only sampling variance, and with one trained manager per configuration, training variance is not measured.

%% file: sections/60ethical.tex
All datasets are public and contain only what users posted publicly.
Because \method writes natural-language descriptions of users, applying it to private data could reveal sensitive attributes.
The Buffer stores unverified inferences. Although the downstream task model never reads it, it should follow the same access, retention, and deletion policies as the Profile.
Holding single-interaction inferences in the Buffer may also reduce stereotyping from sparse evidence, although this is not measured here.
Deployments should let users view and correct their natural-language Profile.
Future interactions are used only during training and are never available at deployment.
We use outputs of GPT-6.1-sol only as training targets for research, name the models used in Sections~\ref{sec:experiments-metrics} and~\ref{sec:experiments-implementation-details} and Appendix~\ref{app:implementation-details-judge}, and release only the code, not the trained manager weights.

%% file: sections/73implementation.tex
\subsection{Training and Targets}
\label{app:implementation-details-training}
\label{app:implementation-details-targets}
Each stage fine-tunes a LoRA~\cite{hu2021lora} adapter on Qwen3.5-4B: \studentdistill and \teacherdistill train separate adapters, and POPD continues the Student adapter with the merged Teacher fixed.
Each stage trains for at most two epochs, evaluates the validation users every three steps, stops early when the validation error has not improved for three consecutive evaluations, and keeps the best validation checkpoint.
The hyperparameters, including the POPD weight $\alpha$, are fixed once and used for all three datasets.
The manager samples at temperature 1 with reasoning off, and its Buffer, Rulebook, and output each have a budget of 1,024 tokens. Dual Specialization uses LoRA rank 32, scaling 64, and dropout 0.05, a learning rate of $10^{-4}$ with 3\% warmup and cosine decay to a tenth, and batch size 16. POPD uses a learning rate of $10^{-5}$ with 16 users per step, reverse KL, and $\alpha=0.5$. The downstream task model uses LoRA rank 32, a learning rate of $2\times10^{-5}$, batch size 32, at most eight epochs with validation every 0.125 epochs and early stopping with patience 3, and inputs of at most 16,384 tokens.

GPT-6.1-sol writes the targets of both roles from the current management input, plus the remaining interactions for \teacherdistill, and is told not to write facts that only the future reveals.
After a proposal scoring worse than \textit{keep} under Eq.~\eqref{eq:reward}, up to three more are generated, and the first eligible one is retained.
On Amazon Books and LaMP-QA, \teacherdistill targets that name a future item title (at most 0.2\%) are dropped. LaMP-2 is excluded because its items are movie descriptions without titles.

On Amazon Books, including \textit{keep} targets in Dual Specialization makes \teacherdistill leave the Profile unchanged in 29.9\% of the validation states, against 3.7\% without them, with 26.0\% of its outputs empty, and raises its validation MAE and RMSE from 0.5766 and 0.9140 to 0.5945 and 0.9310. The Student trained through POPD with this Teacher reaches 0.5948 MAE and 0.9245 RMSE on the test users, against 0.5801 and 0.9152 for \method.

$\mathcal L_{\mathrm{KL}}$ uses only the eligible positions $\mathcal I_t$, the tokens of the Student's action list where the output grammar allows more than one token, such as the operation, the target component, an entry reference, or the entry text.
Tokens fixed by the output format, the stop token, and the tokens of an action rejected for exceeding a component budget are excluded from $\mathcal L_{\mathrm{KL}}$, while $\mathcal L_{\mathrm{task}}$ keeps every sampled token.
An output truncated at the length limit is excluded from both terms, and both terms divide their token sums by $D=1{,}024$.
For Table~\ref{tab:divergence}, forward KL reverses the two distributions in Eq.~\eqref{eq:kl} to $\mathrm{KL}(\pi_{\mathrm T}\,\|\,\pi_\theta)$, which weights each token by the Teacher's probability, and JSD compares both distributions with their mixture, so it is symmetric and bounded, over the same positions $\mathcal I_t$.
Unlike \teacherdistill acting as a manager in Table~\ref{tab:signal}, the Teacher in POPD writes no action list and only gives next-token distributions on the Student's own decisions, so its future information reaches the Student only through which of those decisions it favors.

In the POPD runs that produce the evaluated managers, $\mu_{\mathcal B}$ ranges from $-0.03$ to $-0.16$ in every batch on all three datasets while $r_t$ averages within $\pm 0.012$.
Skip penalties account for 12\% to 19\% of the variance of $R_t$ on Amazon Books but 75\% to 90\% on LaMP-2 and LaMP-QA.
On LaMP-2 and LaMP-QA, where each turn has one interaction, the task term thus mainly keeps actions valid and applicable, and Teacher guidance carries most of the content signal, as task reward alone ($\alpha=1$) does not improve on \studentdistill on LaMP-QA (Tables~\ref{tab:popd} and~\ref{tab:signal}).

\subsection{Task Reward and Task Model}
\label{app:implementation-details-task-reward}
\label{app:implementation-details-downstream-task-model}
The task loss $J$ in Eq.~\eqref{eq:reward} is MAE plus RMSE for rating, the 0--1 tag error averaged over the chunk for tagging, and one minus the aspect-based score for answers.
An output that violates the action format receives $-1$, and a nonempty action list with $n_{\mathrm{skip}}$ of its $n_{\mathrm{act}}$ actions unapplied receives an additional $-n_{\mathrm{skip}}/n_{\mathrm{act}}$.
For each method, the downstream task model used for evaluation is fine-tuned on the DTM users, whose inputs carry the Profiles that this method writes for them, so its training data are disjoint from those of Dual Specialization and POPD.
Likewise, the reward task model is fine-tuned per method on the Profiles its models write for the DS users (DTM users on LaMP-QA), not the POPD users it scores.
All task models use Qwen3.5-4B with a LoRA adapter per method and dataset, and for rating they predict the expected rating,
\begin{equation}
\hat{y} = \sum_{k=1}^{5} k\, p(k), \qquad p(k) = \frac{\exp(\ell_k)}{\sum_{k'=1}^{5}\exp(\ell_{k'})},
\label{eq:expected-rating}
\end{equation}
where $\ell_k$ is the logit of rating $k$ (probability 0 outside the top 20 log-probabilities). For tagging, the 15 tags are scored by log-probability.

\subsection{Cost Analysis}
\label{app:implementation-details-compute}
\begin{table}[h!]
\centering
\footnotesize
\caption{Manager cost per turn on the test streams.}
\setlength{\tabcolsep}{2pt}
\begin{tabular}{l|c|cc|cc|cc}
\toprule
\multirow{2.5}{*}{\textbf{Method}} & \multirow{2.5}{*}{\textbf{Calls}} & \multicolumn{2}{c|}{\textbf{Amazon Books}} & \multicolumn{2}{c|}{\textbf{LaMP-2}} & \multicolumn{2}{c}{\textbf{LaMP-QA}} \\
\cmidrule(lr){3-4} \cmidrule(lr){5-6} \cmidrule(lr){7-8}
 & & In & Out & In & Out & In & Out \\
\midrule
Dynamic Profile Prompting & 1 & 466 & 186 & 353 & 74 & 1,316 & 353 \\
TRIPLE & 5 & 2,676 & 350 & 1,919 & 291 & 6,873 & 339 \\
MemGPT & 5.6--6.3 & 20,227 & 980 & 18,062 & 849 & 21,139 & 1,096 \\
\method & 1 & 2,702 & 308 & 2,817 & 185 & 3,337 & 114 \\
\bottomrule
\end{tabular}
\label{tab:cost}
\end{table}
Table~\ref{tab:cost} compares the input and output tokens that each profile manager spends per management turn on the test streams, summed over all of its LLM calls in the turn and counted with the Qwen3.5-4B tokenizer. Because every manager runs on the same Qwen3.5-4B backbone, cost is compared in tokens and LLM calls per turn.
Table~\ref{tab:cost} averages over all management turns of the test streams (1,000 for Amazon Books, 1,837 for LaMP-2, and 1,000 for LaMP-QA).
Dynamic Profile Prompting and \method make one call per turn, TRIPLE makes one call per candidate refinement, five in total, and MemGPT makes about six calls per turn (5.6 to 6.3 across datasets, including retries) to edit its core memory.
Calls to the downstream task model are not counted.
\method reads 2.5 to 8 times as many input tokens as Dynamic Profile Prompting, since its input adds the Buffer, the Rulebook, and a longer prompt, but on LaMP-QA it writes about a third as many output tokens, since it edits entries instead of rewriting the whole Profile.

\subsection{LaMP-QA Judge}
\label{app:implementation-details-judge}
LaMP-QA is evaluated with the benchmark's aspect-based protocol~\cite{salemi-zamani-2025-lamp}, which extracts from each question the aspects a good answer should address and rates each from 0 to 2, with only the models replaced: GPT-6-luna extracts the aspects from the question body, and Claude Haiku 5.5 (thinking disabled) rates them without seeing the accepted answer.
The training reward takes the per-aspect median of three judgments, and testing uses one valid judgment with up to three retries.
The judge returns no parseable rating for fewer than 0.2\% of answers, which are excluded consistently across methods.
Re-judging with GPT-5.6-terra gives a similar method ranking (Kendall's $\tau_b=0.73$, Spearman's $\rho=0.90$).

%% file: sections/74baselines.tex
All baselines share \method's test users, chunks, evaluation turns, and, except Full History, Profile budget (128 tokens on Amazon Books and LaMP-2 and 256 on LaMP-QA), and each baseline has its own downstream task model, fine-tuned as for \method.
Compared with their original settings, every baseline uses Qwen3.5-4B with reasoning disabled and outputs sampled at temperature 1, except the Stronger Model, and the trained baselines use the Dual Specialization and POPD subsets of \method, without the users of the downstream task model.
These subsets are given in Table~\ref{tab:dataset}, and every trained baseline is thus trained on far less data than in its original setting, which uses many more users and training samples.
\begin{itemize}[leftmargin=1em,itemsep=1pt,topsep=2pt]
\item \textbf{History inputs.} No Personalization gives only the current query, and Recent History gives the most recent interactions that fit within the Profile budget.
Full History gives the whole available history, removing the oldest interactions first when the input exceeds 16,384 tokens, and does not follow the Profile budget.
\item \textbf{Static and Dynamic Profile Prompting.} Static Profile Prompting prompts the backbone once to summarize the initialization interactions into a Profile and keeps it fixed, and Dynamic Profile Prompting prompts it after each chunk to rewrite the previous Profile from the new interactions.
\item \textbf{PURE~\cite{bang2025pure}.} Run by prompting, as in the original method: after each chunk, it extracts likes, dislikes, and key features and appends them to the end of the Profile.
\item \textbf{TRIPLE~\cite{noh2025triple}.} Run by prompting, as in the original method: after each chunk, it writes five refinements and keeps the candidate or the previous Profile with the lowest error on the last chunk:
MAE plus RMSE on Amazon Books, the 0--1 tag error on LaMP-2, and one minus the aspect-based score on LaMP-QA.
\item \textbf{RLPF~\cite{wu2025rlpf}.} As in the original method, trains a one-time Profile writer by REINFORCE~\cite{williams1992reinforce} with a baseline, using the next-turn performance of the downstream task model as the reward.
\item \textbf{LettinGo~\cite{wangl2025lettingo}.} Trains a one-time Profile writer with DPO~\cite{rafailov2023dpo}, as in the original method, on pairs of candidate Profiles ranked by the performance of the downstream task model.
\item \textbf{DEEPER~\cite{chen-etal-2025-deeper}.} Trained with DPO, as in the original method, to refine the Profile after each chunk from prediction-error changes over the previous, current, and next interaction windows.
Training starts from an empty Profile, samples candidates at temperature 1, and gives an invalid candidate a reward of 0, and the error is the 0--1 tag error on LaMP-2 and one minus the aspect-based score on LaMP-QA, as in the task reward.
\item \textbf{Stronger Model.} Runs \method's management prompt and state with GPT-6.1-sol (low reasoning effort, no tools, at most 1,024 output tokens), without fine-tuning.
\item \textbf{MemGPT~\cite{packer2023memgpt}.} Run by prompting with its core-memory tools only: a single core-memory block, limited to the Profile budget, serves as the Profile and is edited with the original append and replace operations in each chunk.
Its archival and recall memory is disabled, since no compared method retrieves past interactions, and its later full-rewrite tool is not used, since it would reduce the method to Dynamic Profile Prompting.
\item \textbf{ACE~\cite{zhang2026ace}.} Keeps its original Reflector and Curator prompts, playbook format, helpful and harmful counters, and duplicate merging, without retrieval.
The downstream task model takes the role of its Generator, the playbook is updated online once per turn without offline epochs, and the Reflector receives the current chunk, the predictions, the observed responses, and the playbook, plus the judge score on LaMP-QA.
Only the entry text of the playbook is given to the downstream task model, and because the original playbook grows without a cap, entries with the lowest helpful-minus-harmful count are removed, the oldest first among ties, until it fits the Profile budget.
\item \textbf{MemAgent~\cite{yu2026memagent}.} Trained with Multi-Conv DAPO~\cite{yu2025dapo}, as in the original method, to overwrite a fixed-size memory after each chunk, with one outcome reward of a stream shared across all memory updates.
The original agent answers a given question over a long document, so, following PersonaMem-v2~\cite{jiang2025personamem-v2}, the update prompt drops the question and asks the agent to keep what is needed to predict the user's next responses.
The memory is limited to the Profile budget instead of the original 1,024 tokens, and training starts without supervised fine-tuning.
\end{itemize}

%% file: sections/75datasets.tex
The three datasets cover a regression, a classification, and a generation task, and an instance in Table~\ref{tab:dataset} is one interaction.
\textbf{Amazon Books}~\cite{hou2024amazon} (regression): each interaction is a book given by its title and categories with the user's rating from 1 to 5, and the downstream task model predicts the rating from the title and categories alone, so it must rely on what it knows about the user.
\textbf{LaMP-2}~\cite{salemi2023lamp} (classification): each interaction is a movie description with the tag the user assigned among 15 tags, the downstream task model predicts the tag, and the official time-based split places a user's dated past items before the training and development questions.
\textbf{LaMP-QA}~\cite{salemi-zamani-2025-lamp} (generation): each interaction is a StackExchange question with its title, body, and accepted answer, and the downstream task model answers a new question from its title alone, as in the original benchmark, where the body states the needs on which the answer is judged.
The original benchmark evaluates one final answer for a given user history, so the streams are rebuilt from the same SE-PQA~\cite{kasela2024sepqa} StackExchange data with the same construction procedure, ordering each asker's questions and accepted answers by time and evaluating answers at successive turns.
The official split, also from SE-PQA, is not used because its validation and test questions overlap the training users.
Users are sampled at random among those with enough history for the full stream.

%% file: sections/76ambiguous.tex
An ambiguous event is an interaction whose interpretation is uncertain when it arrives, because it can reflect a lasting change, a one-time exception, or a preference limited to a specific kind of item.
Aggregate results do not show directly whether \method handles the ambiguous evidence it is designed for, so Section~\ref{sec:experiments-uncertainty-handling-analysis} evaluates the predictions right after such events.
On Amazon Books, it is a rating of 3 or lower in a first-level Books category in which the user has at least two earlier ratings averaging 4 or higher, which can signal a lasting drop in interest or a dislike of one book.
On LaMP-2, it is a gold tag new to the user, and on LaMP-QA, a question in a community new to the user, either of which can mark a new interest or a one-time visit.
Events are detected from earlier interactions only, and the five interactions starting from the first management turn after each event, after the manager has processed it, are evaluated, keeping only those in the same category on Amazon Books.
The tests contain 340 events from 45 users (Amazon Books), 298 from 67 (LaMP-2), and 91 from 36 (LaMP-QA), giving 523, 1,315, and 158 scored event--follow-up pairs (a follow-up counted once per event).
LaMP-QA has fewer scored follow-ups per event than LaMP-2 (1.7 against 4.4), partly because 27\% of its events fall in the last five turns of a stream, where fewer follow-ups remain.

%% file: sections/77analyses.tex
\textbf{Rulebook swap.} For the swap in Section~\ref{sec:experiments-buffer-rulebook-usage}, each test user's Rulebook is replaced at every turn by that of a randomly chosen test user at the same turn, while the rest of the state and the manager stay unchanged. Against each user's own Rulebook, MAE rises from 0.5801 to 0.6020 and RMSE from 0.9152 to 0.9234 on Amazon Books, accuracy falls from 0.5313 to 0.5193 and Macro-F1 from 0.4742 to 0.4541 on LaMP-2, and the LaMP-QA score from 0.4163 to 0.4003. When a rule does not fit, the manager still acts on the current interactions and feedback, which limits the effect of a mismatch.

\noindent\textbf{Rulebook similarity.} Each test user's Rulebook is taken at a quarter, half, and three quarters of the stream and at its end, and embedded with all-MiniLM-L6-v2~\cite{reimers2019sentencebert}. It is compared by cosine similarity with the same user's Rulebook at the next point and with all other test users' Rulebooks at the same point, averaged over pairs. A user's own Rulebooks are more similar (0.95, 0.81, and 0.96 on Amazon Books, LaMP-2, and LaMP-QA) than different users' (0.73, 0.62, and 0.43), which stay fairly high because all Rulebooks share the action format and task vocabulary.

{\emergencystretch=2em \noindent\textbf{Notepad variant.} The notepad variant in Section~\ref{sec:experiments-state-component-ablation} tests whether the gain of the full state comes from the structure of the Buffer and the Rulebook rather than from their capacity. It replaces the two with one unstructured notepad of 2,048 tokens, their combined budget. The manager edits it with add, edit, and delete actions that cite supporting interactions, with no promote action. Only the manager reads it and decides what to keep there, whether hypotheses, rules, or other notes. The variant thus keeps the capacity and edit operations of the full state but not its roles.\par}